**Developing and Evaluating an AI-Assisted Prediction Model for Unplanned Intensive Care Admissions following Elective Neurosurgery using Natural Language Processing within an Electronic Healthcare Record System**

Julia Ive[1*], Olatomiwa Olukoya[2,3*], Jonathan P. Funnell[4], James Booker[4], Sze H M Lam[2], Ugan Reddy[5], Kawsar Noor [1,6,7], Richard JB Dobson[1,6,7,8], Astri M.V. Luoma[5**], Hani J Marcus[2,3,4**]

[1]UCL Institute of Health Informatics, University College London, London, UK

[2]Department of Neurosurgery, National Hospital for Neurology and Neurosurgery, London, UK

[3]UCL Queen Square Institute of Neurology, University College London, London, UK

[4] UCL Hawkes Institute, University College London, London, UK

[5]Department of Neuroanaesthesia and Neurocritical Care, National Hospital for Neurology and Neurosurgery, London, UK

[6]National Institute for Health and Care Research Biomedical Research Centre, University College London, London, UK.

[7]Health Data Research UK, London, UK.

[8]Department of Biostatistics and Health Informatics, King's College, London, London, UK

*equal contribution and joint first authorship

**equal contribution and joint senior authorship

**Corresponding Author:** Dr Julia Ive, email: j.ive@ucl.ac.uk

**Abstract**

Introduction

National guidelines recommend critical care admissions for those surgical patients with a significant 30-day mortality risk or those deemed high risk due to age or comorbidities. Timely care in a specialised neuro-intensive therapy unit (ITU) reduces mortality and hospital stays, with planned admissions being safer than unplanned ones. However, post-operative care decisions remain subjective. This study used artificial intelligence (AI), specifically natural language processing (NLP) to analyse electronic health records (EHRs) and predict ITU admissions for elective surgery patients. The primary goal was to evaluate the effectiveness of AI assistance in minimising decision errors and reducing ITU admission costs.

Methods

This study analysed the EHRs of elective neurosurgery patients from University College London Hospital (UCLH) using NLP. Patients were categorised into planned high dependency unit (HDU) or ITU admission; unplanned HDU or ITU admission; or ward / overnight recovery (ONR). Levels of care were defined according to literature consensus - critical care was defined as level 2 or level 3 care, with level 2 care corresponding to HDU care and level 3 corresponding to ITU care; level 1 care was non-critical care. The Medical Concept Annotation Tool (MedCAT) – a named-entity-recognition (NER) machine learning model available through the CogStack platform – was used to identify Systematised Nomenclature of Medical Clinical Terms (SNOMED-CT) concepts within the clinical notes. We then explored the utility of these identified concepts for a range of AI algorithms trained to predict ITU admission. Finally, we evaluated the results using both performance metrics and interpretability algorithms to identify the key concepts influencing AI predictions. Additionally, we assessed the model's calibration and fairness. We adhered to the TRIPOD+AI guidance for reporting clinical prediction models.

Results

The CogStack-MedCAT NLP model, initially trained on hospital-wide EHRs, underwent two refinements: first with data from patients with Normal Pressure Hydrocephalus (NPH) and then with data from Vestibular Schwannoma (VS) patients, achieving a concept detection F1-score of 0.93. This refined model was then used to extract concepts from EHR notes of 2,268 eligible neurosurgical patients, categorised into planned HDU or ITU admission (349), unplanned HDU or ITU admission (87), and level 1/ward care (1,832).

We integrated the extracted concepts into AI models, including a decision tree model and a neural time-series model. Using the simpler decision tree model, we achieved a recall of 0.87 (CI 0.82 - 0.91) for ITU admissions, reducing the proportion of unplanned ITU cases missed by human experts from 36% to 4%. Our interpretability analysis confirmed that this high performance utilises clinically relevant neurovascular, neuro-oncology, and musculoskeletal concepts in the notes.

Conclusion

This exploratory study successfully demonstrates the potential of AI and NLP to analyse the EHRs of post-operative neurosurgical patients and characterise them based on their post-operative destinations (ward or HDU or ITU care). The NLP model, refined for accuracy, has proven its efficiency in extracting relevant concepts, providing a reliable basis for predictive AI models to use in clinically valid applications. The study highlights the opportunity for AI to aid in allocating resources for neurosurgical patients but requires further research and integration into clinical practice.

## Introduction

Intensive therapy units (ITU) have been developed to provide more detailed observations and invasive treatments for patients who typically require single or multiple organ support (1,2). National guidelines in the USA, UK and elsewhere recommend that routine critical care admissions be common practice for surgical patients with a significant 30-day mortality, or who are deemed to be at a higher risk due to their age and / or comorbidities (1,3,4), to improve patient outcomes. Critical care has been defined in the literature as those requiring level 2 or 3 care, with level 3 corresponding to ITU care where, either advanced respiratory or multi-organ support is required; and level 2 care corresponding to High Dependency Unit (HDU) care reserved for more detailed observations or single organ support (1). Critical care admission is prevalent in neurosurgery due to the high morbidity and mortality that exists. In addition, neurosurgical interventions have significant complications with a 22.2% complication rate following cranial surgery and 11.1% following spinal surgery; 4.8% mortality rate following cranial surgery and 0.5% following spinal surgery (5).

It is well-recognised that critically ill neurological and neurosurgical patients benefit from timely care in a specialised neuro-ITU in the form of reduced mortality and length of hospital stay (6,7). Furthermore, there is evidence that outcomes are better for planned ITU admissions following surgery (1,4,8). Conversely, unplanned intensive care admissions, when patients are triaged pre-operatively as not requiring ITU, but later do, are more costly and have an increased risk of longer hospital stays, morbidity and mortality (10–12). Additionally, due to the aging neurosurgical patient population, the consequences of getting the decision wrong are greater (13–15).

However, there is great variability across centres in how patients are selected for intensive care post-operatively following elective craniotomy (16). It used to be common practice that all patients required intensive care management or monitoring following elective craniotomy (17). However, with the development of post-anaesthesia care units (PACU), the adequacy of HDU, and routine post-operative extubation in the operating room enabling direct ward discharges, there has been a push towards discouraging ITU admission as routine practice (16,17). It remains the case that decisions on the post-operative destination for elective neurosurgical patients are based on expert opinion rather than on robust objective data (17,18). Due to factors such as recency bias, and other human factor error, this decision-making process is far more prone to error with 14 – 28% of ITU admissions being unplanned (12,19,20).

The widespread adoption of electronic health records (EHRs) has led to a significant increase in the exploration of artificial intelligence (AI) to streamline healthcare processes. Among the most pertinent AI subfields in this context are machine learning (ML) and natural language processing (NLP) (6). ML empowers computers to learn autonomously and make predictions by recognising patterns in new data, without the need for explicit programming. This capability is especially crucial for the large-scale analysis of narrative data found in EHRs, where NLP enables computers to extract meaningful information from unstructured text entries.

Predicting unplanned ITU admissions is understudied, with most studies rely on structured data (21). Our study is the first to demonstrate that an NLP-driven algorithm can accurately detect unplanned ITU admissions using unstructured text from EHRs.

Study Aims and Objectives

The aim of this study is to evaluate the effectiveness of NLP algorithms in analysing clinical information from EHRs to detect unplanned ITU admissions among elective neurosurgical patients. The intended users of this AI assistant are healthcare professionals, including doctors and nurses, to enhance their clinical decisions with data-driven insights.

Our objectives are:

1. Build a reliable NLP model to accurately extract concepts from the free-text notes of patients undergoing elective surgery.

2. Demonstrate the utility of concept extraction by NLP to support human experts in enhancing decision-making for intensive care settings. This includes a proof of concept evaluating the importance of these extracted features for predicting ITU admissions, particularly unplanned cases.

3. Conduct a thorough interpretability study to confirm the clinical intuition behind the importance scores of the extracted features.

**Methods**

Ethical Considerations

The study was approved by the University College London Hospital (UCLH) Information Governance Board. The study was not registered. No study protocol was prepared for this study.

Study Design

This study analyses the clinical records of elective neurosurgical patients at UCLH using NLP. Written EHRs, including clinical notes, correspondence, and radiology reports, were collected for patients undergoing neurosurgery from April 2019 to September 2021. Patients were categorised into three groups: planned HDU or ITU admissions, unplanned HDU or ITU admissions, and all others.

Planned admissions were defined as post-operative destinations that were pre-booked and agreed upon before the surgery. Unplanned admissions were defined as either: a) a decision to admit to neurocritical care after the commencement of surgery, or b) admission to ITU or HDU up to 14 days post-surgery due to complications (22).

Patients were identified from a prospectively maintained clinical database. According to the consensus statement released by the Intensive Care Society (23), ITU corresponds to level 3 care, HDU to level 2 care, and "everyone else" to level 1 or ward-level care.

Eligibility Criteria

We included all patients who underwent elective neurosurgical procedures at our hospital between April 2019 and September 2021. Exclusions were made for patients undergoing interventional radiological procedures, those whose electronic health records were unavailable, and patients undergoing carpal tunnel or ulnar decompression, nerve and muscle biopsies, or the insertion of sacral nerve modulators. Patients whose pre-operative urgency of surgical interventions could not be determined according to the NCEPOD (National Confidential Enquiry into Patient Outcome and Death) (24) classification or prior post-operative decisions were excluded. To ensure unique entries across the cohorts, only the first operation for each patient was included.

Cohort Characteristics

From the UCLH databases, 2,268 patients met the eligibility criteria for inclusion. Among these, 349 were admitted to HDU or ITU in a planned manner, 87 were admitted in an unplanned manner, and 1,832 required either level 1 or ward-level care post-operatively.

Demographic information, including age, sex, and ethnicity, was extracted from structured data elements within the EHR. Additionally, diagnoses, signs, symptoms, and co-morbidities were also recorded.

The overall median age at the time of operation was 56 years (ranging from 16 to 95 years), with a male to female ratio of 1:1.05 (49% males, 51% females). Most patients identified ethnically as White (60%).

Natural Language Processing (NLP) Model: Training and Evaluation

We used the CogStack information retrieval platform to extract written clinical information for eligible patients from the EHR system (Epic Systems, Verona, USA) (25). This extracted information was then processed through the CogStack Natural Language Processing platform, enabling analysis with the Medical Concept Annotation Tool (MedCAT), a named-entity-recognition (NER) machine learning model, which identified Systematized Nomenclature of Medicine Clinical Terms (SNOMED-CT) concepts within the clinical notes (25,26).

The MedCAT model was trained in two distinct phases. The first phase involved unsupervised learning to develop a base model using non-specific medical records. During this phase, we pre-trained the model on 1 million randomly sampled EHRs from the hospital-wide database. Following this, the model was fine-tuned using 500 randomly sampled documents annotated with SNOMED-CT concepts by two independent assessors, creating a versatile base model.

In the second phase, this base model underwent further fine-tuning for specific conditions. Initially, it was fine-tuned for Normal Pressure Hydrocephalus (NPH), and subsequently for Vestibular Schwannoma (VS). This two-step fine-tuning process resulted in a final model with enhanced applicability to elective neurosurgical patients.

During this second stage the base model pre-annotated 300 documents from NPH patients. These annotations were then validated in a blinded manner by two independent assessors using the MedCATTrainer interface (27). The assessors reviewed the model's annotations to label terms as correct or incorrect or selected a different concept that best represented the written information. Discrepancies between the assessors were resolved through discussion or adjudication by a third assessor if necessary (see (28) for details).

This first refinement phase achieved a high inter-annotator agreement of 97%. The validated dataset was then used to train the ML model, utilising 80% of the eligible documents. The remaining 20% of the documents were reserved for testing, and model accuracy was assessed using cross-validation on the test set. Cross-validation on the test set involves dividing the test set into *k* subsets and evaluating the model *k* times. This approach provides a more reliable estimate of model performance, reducing the impact of any one test split that might be unusually easy or difficult. The refined model demonstrated high precision and recall, achieving a macro F1-score of 0.92 across identified SNOMED-CT concepts. Precision measures the proportion of examples that the model correctly identifies as relevant out of all the cases it predicts as relevant. Recall measures the proportion of actual relevant examples that the model identifies. The F1-score combines precision and recall into their harmonic mean, providing a balanced measure of the model's performance.

The second refinement followed a similar process using records from VS patients. The model annotated 300 documents from the VS patient dataset, which were validated by two independent assessors using the MedCATTrainer interface. The model achieved an inter-annotator agreement of 72%, indicating substantial agreement. 80% of the validated dataset was used to train the model, while the remaining 20% served as the test set. The model achieved a macro F1-score of 0.93 on the test set in a cross-validation setup (29).

This twice refined and validated base model was then applied to documents from ITU patients meeting the eligibility criteria. Due to time constraints, validation of the model's performance on these records was not possible. However, given the overlap in patient groups, we expect assessment results to be similar to the ones reported above.

Meta-annotations were employed to enhance the accuracy and relevance of the extracted information for our ITU cohort. These annotations are particularly valuable in cases where, despite accurately capturing a term's intended meaning, contextual qualifiers render it irrelevant to the patient. For instance, a documented diagnosis might be negated, as in "COVID-negative"; it might pertain to another individual, as in "family history"; or it might indicate a suspected diagnosis. Terms marked with meta-annotations such as "Negation = Yes," "Experiencer = Other," and "Certainty = Suspected" were excluded from further analysis, as they were deemed not directly relevant to the patient."

Predictive Models: Training and Evaluation

We considered the task of predicting ITU admissions or ward stays a binary prediction problem. We explored the potential of concepts extracted by CogStack using both time-unaware Random Forests (RFs) (30) and time-series Long Short-Term Memory (LSTM) networks (31). We investigated both time-series and non-time-series approaches to effectively capture the intricacies of patient data. Both RFs and LSTM networks are well-established techniques within the AI community, particularly in healthcare applications.

The RF algorithm constructs multiple decision trees using random subsets of the data and combines their results to enhance accuracy. Each decision tree classifies data by forming a tree-like structure where internal nodes represent decision rules, branches represent outcomes, and leaf nodes represent final predictions. This algorithm does not account for patient timelines and uses the counts of concepts from each patient's notes as inputs. We employed the *K*-Best feature selection method, based on the Chi-squared statistic, to identify the most relevant features for RFs. This method selects the top *k* features that have the highest Chi-squared scores, indicating their importance in distinguishing between different classes.

In contrast, LSTM networks are designed to handle sequential data, processing patient notes one at a time. Following best NLP practices, we encoded patient notes using the BERT (Bidirectional Encoder Representations from Transformers) algorithm (32). BERT, based on transformer models, processes words in relation to all other words in a text span, providing a deep understanding of context and encoding the interdependencies of concepts in clinical notes.

To ensure the robustness of our outcomes, we trained multiple models with different random seeds and employed the cross-validation technique. This technique involves partitioning the dataset into *k* "folds" or subsets (we used *k*=5 in our case). Each model is trained on a combination of these folds and tested on the remaining fold, rotating through all folds to ensure each is used for both training and testing.

We used Scikit-learn (33) and Pytorch (34). The implementation details and hyperparameters for our models are detailed in Section A.1 of the Appendix. Hyperparameters were chosen via manual search. We used the standard threshold of 0.5 as a decision rule to convert model probabilities into classification outcomes. This means that if the predicted probability of an ITU was 0.5 or higher, the model classified it as a positive case. Probabilities below 0.5 were classified as ward admissions.

To assess the quality of ITU admission predictions, we calculated the F1-score, precision, recall and False Negative (FN) ratio at the patient level. The False Negative (FN) ratio refers to the proportion of times a model fails to identify an admission when it is present.

We assessed calibration curves to evaluate how well the predicted probabilities matched actual outcomes. Calibration curves are computed by dividing predicted probabilities into bins and comparing the average predicted probability and actual ITU rate for each bin. Accurate probabilities could help healthcare professionals to adequately assess risk levels. Additionally, we assessed fairness by evaluating classification parity across sex and ethnicity demographic subgroups. We measured the ratios of our performance metrics (precision, recall, etc.) for these subgroups (35). For example, we calculated the ratio of each performance metric for the Male group divided by the Female group, and similarly, we did the same for the White group divided by the Non-white group.

To gain deeper insights into how our models make predictions, we employed SHAP (SHapley Additive exPlanations) (36) values and LIME (Local Interpretable Model-agnostic Explanations) (37) interpretability techniques. SHAP values quantify the overall contribution of each feature to the prediction, indicating whether it positively or negatively impacts the outcome for any given input. This helps us understand the global behaviour of the model. On the other hand, LIME focuses on individual predictions by highlighting the most influential features for a specific patient, thereby elucidating the model's decision-making process for that particular case.

Results

Having outlined our methodological approach, including the use of ML techniques and interpretability tools, we now turn to the results of our analysis. The following section presents the key findings from our study, highlighting the performance metrics of our models and the insights gained from the data.

Demographics

Table 1 provides a detailed summary of the baseline statistics and demographics for ITU admissions and ward stays. Given the small sample size for ITU admissions, we averaged the results across subtypes to ensure a more reliable analysis. The data indicates that the median age for both ITU and ward patients is similar, at 56 and 57 years respectively, with ward patients exhibiting a slightly broader age range (16-95 years). The sex distribution is balanced, with an equal percentage of females in both ITU admissions and ward stays (51%). Ethnicity data shows a slightly higher proportion of White patients in ward stays (62%) compared to ITU admissions (57%).

Note that the notes for 58 patients were unavailable from Epic at the time of extraction.

**Table 1. Patient Demographics and Baseline Statistics for ITU Admissions and Ward Patients.**

|  | Count of patient IDs | Median age, range | Male/Female, % | Ethnicity White / Non-white, % |
|---|---|---|---|---|
| Planned | 319 | 56 (16-87 years) | 45% / 55% | 51% / 49% |
| Unplanned | 79 | 57 (19-82 years) | 54% / 46% | 63% / 38% |
| Ward | 1812 | 56 (16-95 years) | 49% / 51% | 62% / 38% |
| Overall | 2210 | 56 (16-95 years) | 49% / 51% | 60% / 40% |

57,433 documents (47,048 level 1/ ward; 8,687 planned HDU or ITU; 1,698 unplanned HDU or ITU) were extracted from the 2,210 patients. All documents contained codable SNOMED-CT concepts and were predominantly either progress notes, plan of care notes or nursing notes.

Overall, 8,300 unique SNOMED-CT concepts were extracted from the medical records. Of these, 6,462 concepts appeared more than once, with 2,980 unique concepts in planned admissions, 1,369 unique concepts in unplanned admissions, and 5,879 unique concepts in level 1 or ward-level care.

Descriptive Statistics

Before applying ML to the concepts extracted by NLP, we first analysed the statistical differences in concept frequency counts between ITU and ward patients. Due to the limited sample size of unplanned ITU admissions, we did not separate the data into planned and unplanned categories. To address the significant imbalance in patient counts between groups, we analysed normalized frequency counts using the Chi-squared test to determine whether there were significant differences between ITU and ward patients. We selected the top 50 concepts based on their highest and lowest frequency ratios between ward and ITU patients. This selection allowed us to investigate the concepts that are more frequent among ward patients and those that are more common among ITU patients, respectively (see A.2 and A.3 in Appendix).

The results of our analysis show that ward patients exhibit rare neurological disorders, such as trigeminal nerve disorder and moyamoya disease (both not seen in ITU patients). They also frequently have cardiovascular disorders, like aortic valve disorder (not seen in ITU) or aneurysm of the posterior cerebral artery (over 100 times more frequent than in ITU). Endocrine disorders, including central hypothyroidism and steroid-induced hyperglycemia (both not seen in ITU), and gastrointestinal disorders, such as carcinoma of the esophagus (not seen in ITU) and feeding problems (over 100 times more frequent in ward than in ITU), were also very common compared to ITU patients.

In contrast, ITU patients presented more frequently with infectious diseases (22 times more frequent than in ward patients). Pain is another common symptom in ITU (3 times more common). ITU is also often associated with conditions like wounds (4 times more frequent) and musculoskeletal findings (19 times more common), including osteoporosis (15 times more frequent).

These trends capture both chronic and acute comorbidities, with a greater representation of dynamic comorbidities in the ITU cohort. This may reflect the stability/instability divide, where ITU patients are more likely to be frailer and have a lower physiological reserve following an intracranial neurovascular event, making them more susceptible to pain and infections.

Predictive AI Models

For our predictive models in the binary classification scenario, we first filtered the patient notes in the original dataset to include only those within the 30-day window before the operation. This filtering reduced the dataset by 33 patients who did not have any notes in that period (see Table 2). After a manual inspection, no specific reason was identified for the absence of these notes. For training, most samples came from the ward category, resulting in a heavily imbalanced training set with a 10:1 patient ratio for ward patients versus ITU admissions. In contrast, we designed a balanced test set with an equal ratio of ward patients to ITU admissions (1:1) via random sampling. It is important to note that all 75 cases of unplanned admissions were excluded from the training data and included in the test set. Clinicians miss those 36% of actual HDU/ITU cases (i.e., the unplanned cases). The primary goal is to evaluate the effectiveness of automated assistance in minimising decision errors and reducing ITU admission costs.

In the cross-validation setting, our predictive models demonstrated stable performance with relatively low standard deviation across different training folds (see A.4 in Appendix). This stability ensures reliable and consistent prediction outcomes.

**Table 2. Summary of Training and Test Data Statistics for Predictive Models (30-Day Pre-Operation Cutoff for Patient Notes)**

|  | Train set, patients count | Test set, patient count |
|---|---|---|
| Planned ITU | 155 | 135 |
| Unplanned ITU | 0 | 75 |
| Ward | 1611 | 201 |
| Total | 1766 | 411 |

**Table 3. Performance Comparison of RF and BERT+LSTM Models for the ITU Test Set.** Results are averaged over 100 runs for the RF model and 5 runs for the BERT+LSTM model. We report precision, recall, F1-score, and False Negative (FN) ratio for each patient group, with a breakdown of the ITU group into planned and unplanned admissions. The best results are highlighted in bold. To estimate the mean and 95% confidence intervals, we employed bootstrapping by resampling the test examples with replacement 1,000 times.

|  | RFs |  |  |  | BERT+LSTM |  |  |  |
|---|---|---|---|---|---|---|---|---|
|  | Prec ↑ (mean, CI) | Recall ↑ (mean, CI) | F1 ↑ (mean, CI) | FN ↓ (mean, CI) | Prec ↑ (mean, CI) | Recall ↑ (mean, CI) | F1 ↑ (mean, CI) | FN ↓ (mean, CI) |
| Ward | **0.88 (0.84-0.92)** | 1.00 (1.00-1.00) | **0.94 (0.91 - 0.96)** | - | 0.84 (0.80 - 0.89) | 1.00 (1.00-1.00) | 0.92 (0.89 - 0.94) | - |
| ITU | 1.00 (1.00-1.00) | **0.87 (0.82 - 0.91)** | **0.93 (0.90 - 0.95)** | **0.13 (0.09-0.18)** | 1.00 (1.00-1.00) | 0.82 (0.77 - 0.87) | 0.90 (0.87 - 0.93) | 0.18 (0.13 - 0.23) |
| Planned | 1.00 (1.00-1.00) | **0.85 (0.80 - 0.91)** | **0.92 (0.88 - 0.95)** | **0.15 (0.09 - 0.21)** | 1.00 (1.00-1.00) | 0.81 (0.74 - 0.87) | 0.89 (0.85 - 0.93) | 0.19 (0.13 - 0.26) |
| Unplanned | 1.00 (1.00-1.00) | **0.89 (0.82 - 0.96)** | **0.94 (0.90 - 0.98)** | **0.11 (0.04-0.18)** | 1.00 (1.00-1.00) | 0.86 (0.77 - 0.93) | 0.92 (0.87 - 0.96) | 0.14 (0.07 - 0.23) |

**Prediction Results**

Results of our experiments are presented in Table 3. Both RF and time-series BERT+LSTM models achieve high precision (0.93 on average), recall (0.92 on average), and F1-scores (0.92 on average) across classes. In terms of performance per class, we observe notable variations: higher precision (1.0) and lower recall (0.85) is characteristic for the ITU patients but higher recall (1.0) and lower precision (0.86) for the ward patients. This is because ward patients are overrepresented in our training data leading to the models that better capture patterns related to this group.

The table indicates that RFs perform better than BERT+LSTM. The RF model achieves perfect precision and a recall of 0.87 for ITU admissions, 0.85 for planned ITU admissions, and 0.89 for unplanned ITU admissions. This highlights the model's potential to identify cases that clinicians might otherwise overlook. These metrics are, on average, 3 points higher than those for the BERT+LSTM model. The FN rates are 0.13 for ITU, 0.15 for planned, and 0.11 for unplanned admissions, averaging 0.04 lower than for the BERT+LSTM model. Notably, only 11% of unplanned ITU cases remain undetected, significantly reducing the total proportion of ITU cases missed by human experts from 36% to just 4%.

It is important to note that the RF model does not consider the temporal aspect of patient data; it simply aggregates the concept counts from each patient's notes. As described in the Methods section, before inputting concept counts into the RF models, we identified the most important concepts using a feature selection algorithm based on the Chi-squared statistic. We selected the 20 most significant features to build our RF models. The number of features was a hyperparameter we fine-tuned manually (see Figure 2). These selected features align with the outcomes of our previous statistical analysis of concept distributions. In our previous analysis, we observed that neurological conditions, musculoskeletal issues including osteoporosis, pain, and gastrointestinal conditions were important in distinguishing between ward and ITU patients. Our feature selection process identifies neurovascular, neuro-oncological, and infection concepts (e.g., "Intracranial meningioma," "Intracranial Aneurysm," and "Fungal Infection of Central Nervous System"). It also highlights musculoskeletal concepts such as "Blister of Skin and/or Mucosa" and "Osteoporosis," gastrointestinal concepts like "Carcinoma of Esophagus," and pain-related concepts such as "Inguinal Pain."

Additionally, the external ventricular drain (EVD) is an important concept in ITU. It is essential for managing acute conditions that require immediate intervention to control intracranial pressure. In this context, "battery" specifically refers to the VNS (Vagus Nerve Stimulator) battery. The proper functioning of this device is vital for patients with epilepsy or depression, as it plays a crucial role in managing these chronic conditions. "Dry skin" primarily indicates the state of post-operative wounds that heal.

**Figure 2. Top 20 Most Important Concepts Selected by the Chi-Squared Statistic for the RF Algorithm for ITU Admission Prediction.** The selected features align with our statistical analysis, emphasising the significance of neurovascular, neuro-oncological, infection and musculoskeletal concepts for ITU / ward patients. Note that Battery stands for VNS (vagus nerve stimulation) battery, and EVD refers to the external ventricular drain.

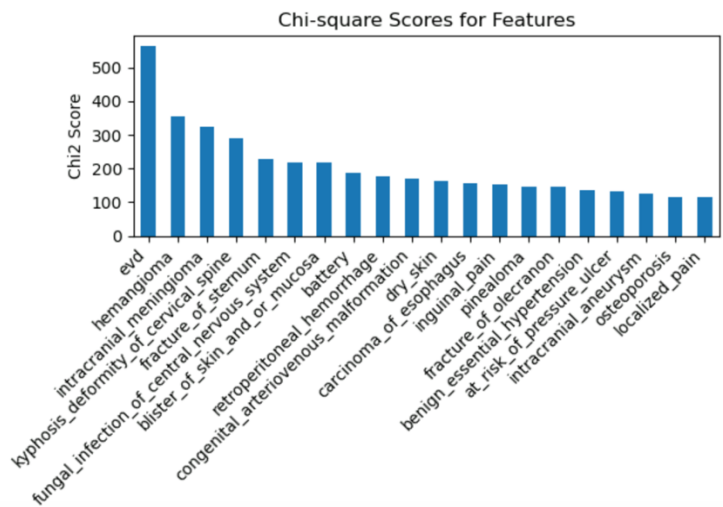

In contrast, BERT+LSTM processes each patient note individually and examines the sequential dependencies of those notes. This approach might explain the increase in missed ITU cases (0.04 higher FN rates than for RFs), as the model's complexity may hinder its ability to learn effectively from the limited data.

We attempted to investigate whether RFs and the BERT+LSTM model identify different cases. To do so, we averaged the output probabilities of both models and applied a decision rule with a threshold of 0.5 to predict ITU admissions. However, we observed that the detected ITU cases overlapped, and this ensembling approach did not provide any improvement over using RFs alone (see A.5 in Appendix).

We further evaluated our best RF model by measuring its calibration and fairness. Among the three models, RFs demonstrated better calibration, remaining closest to the ideal line, especially in mid-range probability values. In contrast, the BERT+LSTM and Ensembling models exhibited stronger miscalibration, underestimating the probability of ITU due to imbalanced data (see A.6 in Appendix).

Our RF model demonstrated classification parity across sex and ethnicity groups (0.99 ratio across metrics for both Male/Female and White/Non-white subgroups). The only notable difference was a slightly lower FN rate for Female and Non-white patients (1.29 ratio on average across demographic groups), suggesting better recall for these subgroups (see A.7 and A.8 in Appendix).

**Explainability**

Our models demonstrate very good predictive capabilities. However, these numerical values alone do not provide any insights on whether the model decision-making procedure is correct. Hence, we conduct a thorough explainability analysis. We use the two popular interpretability techniques: SHAP (Shapley Additive Explanations, (36)) and Local Interpretable Model-agnostic Explanations (LIME, (37)) to understand how individual concepts contribute to model predictions and whether those contributions could be justified from the clinical perspective. In our interpretability analysis we focus on the RF model for simplicity.

SHAP values provide feature importances across all model predictions. This approach directly links back to our feature selection process, enabling us to understand how changes in each concept count influence the predicted probability of ITU admission (see Figure 3 for predictions over the training data). Positive values on the x-axis indicate an increased chance of ITU admission, while negative values reduce this chance. The colour on the y-axis represents concept counts, with red indicating high values and blue indicating low values. From Figure 3, we can see that frequent mentions of osteoporosis and dry skin decrease the chance of ITU admission, while conditions such as intracranial meningioma and aneurysm increase those chances.

Note that the Chi-squared feature importance plot (Figure 2) and the SHAP feature importance plot highlight different aspects of the feature selection process. The Chi-squared test measures the statistical dependence between each feature and the target variable. Features with the highest Chi-squared scores, such as "EVD," "Hemangioma," and "Intracranial meningioma," indicate a significant association with the target label. On the other hand, Shapley values measure the contribution of each feature to a model's prediction, showing how individual features interact with each other and contribute to the model's decision-making process. Here, features like "Osteoporosis" and "Dry skin" appear at the top. Although statistical analysis showed that "Osteoporosis " and "Dry skin" are more frequent for ITU patients (see A.3 in Appendix), their presence in combination with other features steers the model towards predicting a ward stay. All these differences suggest that some features with high statistical correlation may not influence prediction outcomes in the expected way, while others with lower Chi-squared scores can have a higher contribution in combination with other features. Additionally, features with higher Chi-squared scores may influence the model in unexpected ways when considered in complex interactions with other features.

**Figure 3. SHAP Global Feature Importance Outcomes for the RF Model on the ITU Training Set.** The top features for ITU admission are sorted by their mean absolute SHAP importance values. Each row indicates the impact of each concept on predictions, with each data sample represented as a dot. The colour of the dots ranges from blue (lower concept frequency values) to red (higher frequency values). The horizontal position of a dot (SHAP value) shows the concept's contribution to the prediction, with right indicating ITU admission and left indicating ward. For example, in the top row, less frequent mentions of osteoporosis increase the chances of ward stay, while frequent mentions of intracranial meningioma increase the probability of ITU admission.

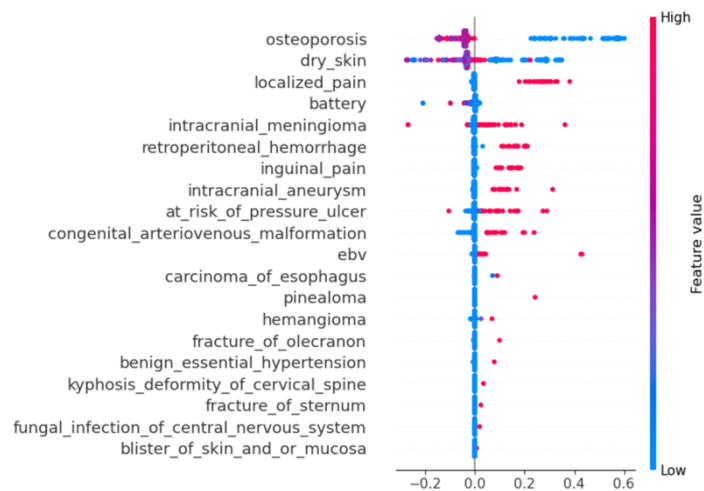

In contrast, the LIME technique offers localised explanations for each individual example (see Figure 4). By applying LIME to ITU admission predictions, we can examine why the model classified one patient as requiring ITU admission while another was not flagged. We have thoroughly investigated our LIME explanations and discovered that correct predictions of unplanned admissions rely on infectious, pain, and musculoskeletal conditions. False negatives, or missed unplanned cases, are frequently associated with mentions of osteoporosis and dry skin.

To conclude, we observe high coherence between model explainability outcomes and statistical observations confirming clinical validity of our high prediction performance.

**Figure 4. LIME Analysis of an Unplanned ITU Case Flagged by our Model (True Positive Prediction).** The x-axis represents concept importance, where positive values (orange) indicate features that contribute to ITU admission, while negative values (blue) indicate concepts that contribute to ward stay. The y-axis lists the contributing concepts and their importance scores. In this example, the neurological disorders "Congenital arteriovenous malformation" and "Intracranial meningioma" contribute to the correct prediction of admission with highest importance scores. The scores of other concepts fall below the threshold of floating-point precision.

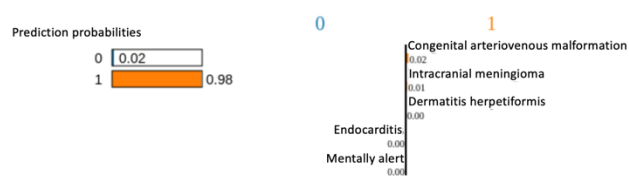

**Discussion**

Principal findings

This exploratory study serves as a proof of concept and demonstrates the ability of an artificial intelligence (AI) natural language processing (NLP) model to analyse the electronic health records (EHRs) of post-operative elective neurosurgical patients and effectively characterise their post-operative destinations.

This is exciting given the novelty and the potential scope of its application. AI has already demonstrated its utility within neurosurgery in various ways, such as in automatic tumour segmentation; the prediction of symptomatic vasospasm following subarachnoid haemorrhage; and in the prediction of survival in patients with Glioma (38,39). Similarly, NLP, has been utilised in the field of spinal surgery to analyse operative notes, radiology reports, and social media to detect outcomes such as post-operative infections, venous thromboembolism, and patient satisfaction (40).

The CogStack NLP model was twice-refined, first using the data of patients with NPH and then with the data of Vestibular Schwannoma patients. The model was proven, following these rounds of refinement, to be both accurate (substantial or high inter-annotator agreement) and precise (F1-score 0.93 on average) (28, 29). Our investigations of the extracted concepts revealed significant differences in the prevalence of infectious, pain and musculoskeletal conditions which were much more common in ITU patients. In contrast, rare neurological disorders, cardiovascular and endocrine disorders were more prevalent in ward patients. This distinction may reflect the greater frailty and lower physiological reserve of ITU patients, making them more susceptible to infections.

We integrated the extracted concepts into two AI predictive models: Random Forests (RFs) and Long Short-Term Memory (LSTM) networks enhanced with BERT embeddings. The simpler and more intuitive RF model was the clear winner, with a recall of 0.89 (CI 0.82 - 0.91) for unplanned ITU, reducing the total proportion of ITU cases missed by human experts from 36% to just 4%. Our thorough interpretability analysis of the RF model, both at the algorithm level and for individual examples, confirmed the importance of neurovascular, neuro-oncological, and musculoskeletal concepts in the model's decisions, thereby validating the clinical relevance of the predictions. Our RF model demonstrated satisfactory calibration and very good classification parity across sex and ethnicity groups.

Our findings provide promising evidence that using similar AI predictive models in clinical practice could improve decision-making, reduce medical errors, and lower hospital admission costs.

Comparisons to the literature

Intensive care units account for around 22% - 33% of the total costs within a hospital and are on average three to five times greater than the cost of ward level care; as such, there has been a greater emphasis on ensuring that care within such units is reserved for those who truly need it. There is, however, currently no standardised criteria for post-operative ITU allocations, with the decision being based on expert opinion and with the current model, unplanned ITU admissions remain as high as 14 – 28% (17–20,41).

Given the above, there has been recognition of the need to characterise these patients, and to do so in a timely manner to improve healthcare outcomes and reduce ITU emergency costs. One such study, by Hanak et al identified that age and diabetes are predictive of the ITU admission on multivariate analysis (41). Similarly, Memtsoudis et al. characterised critical care requirements in patients undergoing lumbar spine fusion and identified that comorbidities such as myocardial infarction, peripheral vascular disease, COPD, uncompleted and complicated diabetes, cancer, and sleep apnoea as well as older age were all significant predictors of the need for ITU level care, with complicated diabetes and sleep apnoea carrying the highest risk (43).

One of the advantages of AI and machine learning (ML) is the ability to learn without direct programming and to recognise patterns in vast quantities of data than can be amassed by a single person (38). There has been a growing awareness of the utility of AI in healthcare: AI has been used to automatically classify epilepsy with 60% accuracy (62% clinician accuracy); predict tumour type, including glioma, with an 86% accuracy; and diagnose cerebral aneurysms with >90% accuracy (38).

Predicting unplanned ITU admissions remains understudied. A related study of Yao et al. (2) developed a prognostic model for unplanned ITU care. Following logistic regression, they identified that the lesion type, a Karnofsky performance status <70 at admission, higher fibrinogen and blood glucose levels at admission, longer duration of surgery, greater intra-operative blood loss, and post-operative surgical bed bleeding on computed tomography were independently predictive of unplanned ITU admissions. These parameters were utilised in their model and following external validation, they demonstrated good discrimination (c-statistic 0.811) and calibration (Hosmer-Lemeshow P = 0.141) for unplanned ITU admissions. This study used only structured tabular data.

However, a significant proportion of the data within health record systems are unstructured free text entries. NLP enables computers to understand human languages and thus make sense of, and extract information from these unstructured entries (39). A platform that may provide a way forward is CogStack, which functions as an NLP information retrieval platform (25).

Our study is the first of its kind to demonstrate the ability of an NLP-driven algorithm to accurately detect unplanned ITU admissions using unstructured text for post-operative elective neurosurgical patients.

Strengths and Limitations

We demonstrated that our NLP model (26) effectively and accurately extracts clinical concepts and generalises well to new hospital settings, including neurosurgical patients. One of the major strengths of this study is the robust final model, which was trained on a large dataset and further refined for optimal performance using data from patients with normal pressure hydrocephalus and subsequently vestibular schwannoma, increasing its applicability to a broad range of such patients. We have proved the utility of the concepts extracted by this model for accurate and transparent automated prediction of ITU admission.

The large dataset is a strength, but the fact that it was obtained retrospectively and from a single centre increases the risk of potential overfitting, despite the demonstrated generalisation capacity.

A further limitation is that, with post-operative destination triage being based on expert opinion, a few patients may have been selected for a planned post-operative HDU or ITU admission who may not have truly needed HDU or ITU level support or intervention whilst there. This might skew the interpretation of the data as the planned HDU or ITU group of patients might be functionally similar to the level 1 or ward category of patients. For this reason, future research should aim to define a true HDU or ITU, whether planned or unplanned, based on whether ITU/HDU intervention was required. A final limitation is the paucity of clinical records prior to the operation date. Analysis of primary care data could similarly be a focus of future research.

When implementing our ITU prediction model, poor quality or missing features should be identified and addressed through data imputation techniques (e.g., *k*-nearest neighbours).

**Conclusions**

Our findings demonstrate the feasibility of developing a natural language processing (NLP) model capable of accurately extracting clinical features from unstructured electronic health records (EHRs) for post-operative neurosurgical patients. In our study, we evaluated the importance of these extracted features and discovered that they are clinically useful, helping to create systems that achieve very high performance in predicting ITU admissions, particularly unplanned cases.

Future research will work towards integrating an AI-assisted "high-risk for HDU or ITU" alert system into the local EHR using CogStack (44).

We adhered to the TRIPOD+AI guidance for reporting clinical prediction models that use regression or machine learning methods (45). The completed checklist is available in the Supplementary Material.

**Code and Data Availability**

The code and data could not be publicly shared due to their confidentiality requirement but can be made available through collaboration. The code was implemented in Python 3.10.

**Acknowledgments**

HJM is supported by the National Institute for Health and Care Research University College London Hospitals Biomedical Research Centre.

**Author contributions statement**

JI: Conceptualisation, Methodology, Software, Validation, Formal analysis, Investigation, Writing – Final draft preparation, Writing - Reviewing and Editing. OO: Conceptualisation, Methodology, Formal Analysis, Investigation, Writing – Early draft preparation, Writing - Reviewing and Editing. KN: Resources, Data Curation. RD, AL, HM: Conceptualisation, Methodology, Writing - Reviewing and Editing. JF, JB, SL, UR: Conceptualisation, Writing - Reviewing and Editing. All authors approved the manuscript.

Competing interests

We have no competing interests to declare.

Appendix

A.1 Algorithm Hyperparameters

| Classifier | Library | Hyperparameters |
|---|---|---|
| **Random Forest** | Scikit-learn (32) | num of estimators=300, criterion="gini", max depth=None, min samples split=2, min samples leaf=1, CV=5, 100 runs |
| **LSTM** | Pytorch (33) | epochs=15, hidden size=128, batch size=4, num of LSTM layers=1, dropout rate=0.5, learning rate=1e-3, CV=5, 5 runs |

A.2 Statistical Analysis of Concept Counts Between ITU and Ward Patients. This table presents top 50 SNOMED-CT concepts with statistically significant ($p < 0.05$) frequency differences in ward and ITU patients, identified using the Chi-squared test. To address the class imbalance, we normalised the frequency counts and selected the top 50 concepts based on the highest frequency ratio between ward and ITU. This indicates that these concepts are **more frequent for ward patients**, with NA signifying that the concept is absent for ITU patients.

| Concepts | Normalised frequency counts for ward patients | Normalised frequency counts for ITU patients | **Frequency ratio** |
|---|---|---|---|
| neoplasm_of_esophagus | 0.0032 | 0 | NA |
| steroid_induced_hyperglycemia | 0.0032 | 0 | NA |
| aortic_valve_disorder | 0.0032 | 0 | NA |
| quadrantanopia_of_left_eye | 0.0032 | 0 | NA |
| moyamoya_disease | 0.00319 | 0 | NA |
| carcinoma_in_situ | 0.0032 | 0 | NA |
| carcinoid_tumor | 0.0032 | 0 | NA |
| central_hypothyroidism | 0.0032 | 0 | NA |
| glasgow_coma_scale_finding | 0.0032 | 0 | NA |
| embolic_stroke | 0.0032 | 0 | NA |
| herniation_under_falx_cerebri | 0.00319 | 0 | NA |
| upshaw_schulman_syndrome | 0.0032 | 0 | NA |
| hemangioma_of_vertebral_column | 0.0032 | 0 | NA |
| lagophthalmos | 0.00319 | 0 | NA |
| hemangioma_of_liver | 0.0032 | 0 | NA |
| carcinoma_of_esophagus | 0.0032 | 0 | NA |
| trigeminal_nerve_disorder | 0.0032 | 0 | NA |
| epidermoid_cyst | 0.00319 | 0 | NA |
| multiple_malignancy | 0.0032 | 0.00001 | 529.53 |
| leptospirosis | 0.0032 | 0.00001 | 529.53 |
| pollution | 0.0032 | 0.00001 | 529.53 |
| aneurysm_of_posterior_cerebral_artery | 0.0032 | 0.00001 | 529.53 |
| mucocele_of_salivary_gland | 0.0032 | 0.00001 | 529.53 |
| feeding_problem | 0.0032 | 0.00001 | 529.53 |
| cerebellopontine_angle_tumor | 0.0032 | 0.00001 | 529.53 |
| langerhans_cell_histiocytosis | 0.0032 | 0.00001 | 529.53 |
| severe_vertigo | 0.0032 | 0.00001 | 529.53 |
| sensation_of_falling | 0.0032 | 0.00001 | 529.53 |
| hypoplasia_of_spleen | 0.0032 | 0.00001 | 529.53 |
| congenital_anomaly_of_lung | 0.0032 | 0.00001 | 529.53 |
| chronic_kidney_disease_stage_5 | 0.0032 | 0.00001 | 529.53 |
| transmitted_sounds | 0.0032 | 0.00001 | 529.53 |
| chronic_periodontitis | 0.0032 | 0.00001 | 529.29 |
| necrosis_caused_by_ionizing_radiation | 0.0032 | 0.00001 | 529.29 |
| myelitis | 0.0032 | 0.00001 | 529.29 |
| superficial_thrombophlebitis | 0.0032 | 0.00001 | 529.29 |
| dental_calculus | 0.0032 | 0.00001 | 529.29 |
| dental_phobia | 0.0032 | 0.00001 | 529.29 |
| peripheral_venous_cannula | 0.0032 | 0.00001 | 529.29 |
| adrenal_adenoma | 0.0032 | 0.00001 | 529.29 |
| exposure_keratoconjunctivitis | 0.0032 | 0.00001 | 529.29 |
| kyphosis_deformity_of_cervical_spine | 0.0032 | 0.00001 | 529.29 |
| infection_caused_by_wuchereria_bancrofti | 0.0032 | 0.00001 | 529.29 |
| discoloration_of_skin | 0.0032 | 0.00001 | 529.29 |
| talipes_equinovarus | 0.0032 | 0.00001 | 529.29 |
| oronary_bypass_graft_finding | 0.0032 | 0.00001 | 529.29 |
| right_inguinal_hernia | 0.0032 | 0.00001 | 529.29 |
| positional_vertigo | 0.00319 | 0.00001 | 529.05 |
| hypertrophic_cardiomyopathy | 0.00319 | 0.00001 | 529.05 |
| myasthenia_gravis | 0.00319 | 0.00001 | 529.05 |

A.3 Statistical Analysis of Concept Counts Between ITU and Ward Patients. This table presents top 50 SNOMED-CT concepts with statistically significant ($p < 0.05$) frequency differences in ward and ITU patients, identified using the Chi-squared test analysis. To address the class imbalance, we normalised the frequency counts and selected the top 50 concepts based on the lowest frequency ratio between ward and ITU. This indicates that these concepts are **more frequent among ITU patients**.

| Concepts | Normalised frequency counts for ward patients | Normalised frequency counts for ITU patients | Frequency ratio |
|---|---|---|---|
| musculoskeletal_finding | 0.00051 | 0.00951 | 0.05 |
| infectious_disease | 0.00021 | 0.00454 | 0.05 |
| osteoporosis | 0.00053 | 0.00811 | 0.06 |
| disability | 0.00029 | 0.00502 | 0.06 |
| digestive_system_finding | 0.00047 | 0.00696 | 0.07 |
| endocrine_finding | 0.00081 | 0.00846 | 0.1 |
| postoperative_nausea_and_vomiting | 0.00104 | 0.00742 | 0.14 |
| pain_in_throat | 0.00101 | 0.00752 | 0.14 |
| allergic_disposition | 0.00081 | 0.00544 | 0.15 |
| normal_respiratory_function | 0.00099 | 0.00626 | 0.16 |
| glasgow_coma_scale | 0.00048 | 0.00274 | 0.17 |
| 15 | 0.00054 | 0.00279 | 0.19 |
| eating_ feeding_drinking_finding | 0.00055 | 0.00291 | 0.19 |
| open_mouth | 0.00137 | 0.00683 | 0.2 |
| oral_intake | 0.00134 | 0.0063 | 0.21 |
| problem | 0.00063 | 0.00296 | 0.21 |
| smoker | 0.00093 | 0.00429 | 0.22 |
| necking | 0.00133 | 0.00566 | 0.23 |
| finding | 0.00073 | 0.00317 | 0.23 |
| well_in_self | 0.00086 | 0.00375 | 0.23 |
| exsanguination | 0.00144 | 0.00637 | 0.23 |
| pressure | 0.00102 | 0.00424 | 0.24 |
| wound | 0.00087 | 0.0037 | 0.24 |
| face_goes_red | 0.00147 | 0.00592 | 0.25 |
| bleeding | 0.00112 | 0.00427 | 0.26 |
| spontaneous_rupture_of_membranes | 0.00155 | 0.00606 | 0.26 |
| nausea | 0.0012 | 0.00438 | 0.27 |
| pupils_equal_and_reacting_to_light | 0.00112 | 0.00404 | 0.28 |
| pain | 0.00108 | 0.00345 | 0.31 |
| backache | 0.00039 | 0.0012 | 0.32 |
| dry_skin | 0.00067 | 0.00208 | 0.32 |
| does_mobilize | 0.00094 | 0.00299 | 0.32 |
| mentally alert | 0.00112 | 0.00356 | 0.32 |
| hematoma | 0.00153 | 0.00444 | 0.34 |
| alcoholic_beverage_intake | 0.00155 | 0.0045 | 0.34 |
| snomed_ct_to_meddra_simple_map_reference_set | 0.00142 | 0.0038 | 0.37 |
| vital_signs_finding | 0.00128 | 0.00343 | 0.37 |
| supine_body_position | 0.00171 | 0.00465 | 0.37 |
| deep_breathing | 0.0019 | 0.00496 | 0.38 |
| skin_appearance_normal | 0.00031 | 0.00081 | 0.38 |
| chest_clear | 0.00164 | 0.00438 | 0.38 |
| o_e (operational environment) | 0.0016 | 0.00406 | 0.39 |
| at_risk_for_falls | 0.00103 | 0.00262 | 0.39 |
| minute | 0.0015 | 0.00372 | 0.4 |
| chest_pain | 0.00125 | 0.00304 | 0.41 |
| breathing_room_air | 0.00153 | 0.00368 | 0.42 |
| discharged_from_hospital | 0.00151 | 0.00307 | 0.49 |
| unconscious | 0.00214 | 0.00424 | 0.5 |
| recommended | 0.00154 | 0.00307 | 0.5 |
| orientated | 0.00162 | 0.0032 | 0.51 |

A.4. Cross-Validation Stability Results: The table reports mean and standard deviation (std) for F1-Score, precision, and recall over the validation set across 5 Folds for each algorithm.

|  | RFs | | | BERT+LSTMs | | |
|---|---|---|---|---|---|---|
|  | Precision, mean/std | Recall, mean/std | F1, mean/std | Precision, mean/std | Recall, mean/std | F1, mean/std |
| Ward | 0.98 / 0.06 | 0.99 / 0.05 | 0.99 / 0.04 | 0.99 / 0.01 | 0.99 / 0.01 | 0.99 / 0.005 |
| Planned ITU | 0.92 / 0.05 | 0.85 / 0.05 | 0.88 / 0.04 | 0.94 / 0.1 | 0.85 / 0.09 | 0.88 / 0.05 |

A.5. Performance of the Ensembling (RFs and BERT+LSTM) Method for the ITU Test Set. This table presents our investigation into whether RFs and our BERT+LSTM model identify different cases. We averaged the output probabilities of both models and applied a decision rule with a threshold of 0.5. There was no improvement over using RFs alone, indicating overlap in the cases detected by both models. To estimate the mean and 95% confidence intervals, we employed bootstrapping by resampling the test examples with replacement 1,000 times.

**Ensemble RFs/BERT+LSTM**

|  | Prec (mean, CI) | Recall (mean, CI) | F1 (mean, CI) | FN Rate (mean, CI) |
|---|---|---|---|---|
| Ward | 0.87 (0.82-0.91) | 1.00 (1.00- 1.00) | 0.93 (0.90 - 0.95) | - |
| ITU | 1.00 (1.00- 1.00) | 0.85 (0.80 - 0.90) | 0.92 (0.89 - 0.95) | 0.15 (0.10 - 0.20) |
| Planned | 1.00 (1.00- 1.00) | 0.84 (0.78- 0.89) | 0.91 (0.88 - 0.94) | 0.16 (0.11 - 0.22) |
| Unplanned | 1.00 (1.00- 1.00) | 0.87 (0.79 - 0.93) | 0.93 (0.88-0.97) | 0.13 (0.07 - 0.21) |

A.6. Calibration Curves. This plot compares the predicted probabilities with the actual outcome probabilities for three models: Ensembling (blue), BERT+LSTM (orange), and RFs (green). The red dashed line represents perfect calibration, where a predicted probability of 70% corresponds to an outcome that occurs 70% of the times. RFs (green) exhibit the best calibration, as their curve remains closest to the perfect line across probability levels.

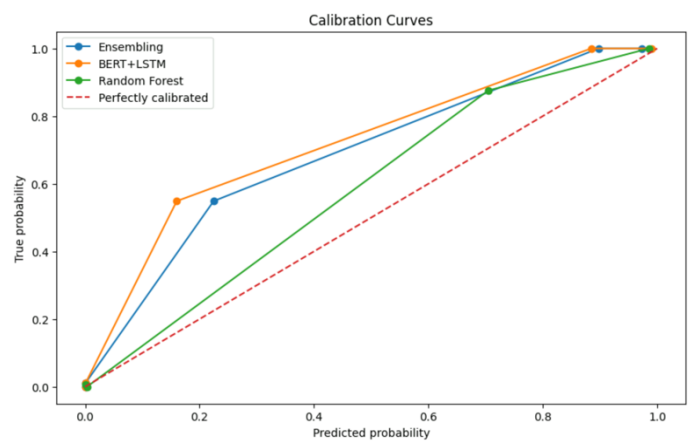

A.7. Classification Parity for Male/Female for RFs. This plot shows the performance metrics and their ratios for Male and Female groups for RFs. A ratio greater than 1 suggests better performance for the Male subgroup, while a ratio less than 1 indicates better performance for the Female group.

**RFs**

|  | Prec Male/Female, Ratio | Recall Male/Female, Ratio | F1 Male/Female, Ratio | FN rate Male/Female, Ratio |
|---|---|---|---|---|
| Ward | 0.86/0.89, 0.97 | 1.00/1.00, 1.00 | 0.93/0.94, 0.99 | NA |
| ITU | 1.00/1.00, 1.00 | 0.86/0.87, 0.98 | 0.92/0.92, 0.99 | 0.14/0.12, 1.14 |

A.8. Classification Parity for White/Non-white for RFs. This plot shows the performance metrics and their ratios for White and Non-white groups for RFs. A ratio greater than 1 suggests better performance for the White subgroup, while a ratio less than 1 indicates better performance for the Non-white group.

|  | RFs | | | |
|---|---|---|---|---|
|  | Prec White/Non-white, Ratio | Recall White/Non-white, Ratio | F1 White/Non-white, Ratio | FN rate White/Non-white, Ratio |
| Ward | 0.88/0.86, 1.03 | 1.00/1.00, 1.00 | 0.94/0.92, 1.01 | NA |
| ITU | 1.00/1.00, 1.00 | 0.85/0.89, 0.95 | 0.92/0.94, 0.97 | 0.15/0.11, 1.43 |